\documentclass[a4paper,11pt]{article}

\usepackage{comment}
\usepackage{graphicx}
\usepackage[clockwise]{rotating}
\usepackage{anyfontsize}
\usepackage{enumerate}
\usepackage{amssymb,amsmath}
\usepackage{soul}
\usepackage{multirow}
\usepackage{algcompatible}
\usepackage{algpseudocode}
\usepackage{algorithm}

\usepackage{pspicture}
\usepackage[pdf]{pstricks}
\usepackage{url}
\usepackage{balance}
\usepackage{todonotes}
\usepackage{refcount}
\usepackage{subfigure}
\usepackage{graphicx}
\usepackage{rotating}
\usepackage{hhline}

\usepackage{authblk}

\begin{document}


\title{Merge Non-Dominated Sorting Algorithm for Many-Objective Optimization}

\author[1]{Javier Moreno \thanks{\texttt{javier.morenom@edu.uah.es}}}
\author[1]{Daniel Rodriguez \thanks{\texttt{daniel.rodriguezg@uah.es}}}
\author[2]{Antonio Nebro C\thanks{\texttt{anotonio.nebro@uma.es}}}
\author[3,4]{Jose~A. Lozano\thanks{\texttt{ja.lozano@ehu.eus, jlozano@bcamath.org}}}
\affil[1]{Dept of Comp Sci, University of Alcala, 28805 Alcala de Henares, Madrid, Spain}
\affil[2]{Dept Lenguajes y Ciencias de la Computaci\'on, Univ of Malaga, 29071 Malaga, Spain}
\affil[3]{Dept of Comp Sci and AI, Univ of the  Basque Country, 20080. San Sebastian-Donostia, Spain}
\affil[4]{Basque Center For Applied Mathematics (BCAM), 48009 Bilbao, Spain}

\renewcommand\Authands{ and }
%


\maketitle

\begin{abstract}
Many Pareto-based multi-objective evolutionary algorithms require to rank the solutions of the population in each iteration according to the dominance principle, what can become a costly operation particularly in the case of dealing with many-objective optimization problems.
In this paper, we present a new efficient algorithm for computing the non-dominated sorting procedure, called Merge Non-Dominated Sorting (MNDS), which has a best computational complexity of $\Theta(NlogN)$ and a worst computational complexity of $\Theta(MN^2)$. 
Our approach is based on the computation of the dominance set of each solution by taking advantage of the characteristics of the merge sort algorithm.
We compare the MNDS against four well-known techniques that can be considered as the state-of-the-art. The results indicate that the MNDS algorithm outperforms the other techniques in terms of number of comparisons as well as the total running time.
\end{abstract}



\section{Introduction}
\label{sec:introduction}

Evolutionary algorithms (EAs) have been successfully applied in the solution of multi-objective optimization problems (MOPs) in the last two decades. These approaches can be mainly classified into Pareto-based, indicator-based and decomposition-based EAs. Most of algorithms belonging to the first group, which includes NSGA-II~\cite{Deb2002}, SPEA2~\cite{Zitzler01spea2:improving} and many others~\cite{BOS}, typically require to rank the population in the selection and replacement phases according to the dominance principle~\cite{Srinivas1994}.

The non-dominated ranking procedure can be computationally significant in the total computing time of a multi-objective evolutionary algorithm (MOEA), particularly when dealing with many-objective problems. 

In this paper, we present the Merge Non-Dominated Sorting (MNDS) algorithm, which is aimed to efficiently perform the non-dominated ranking. MNDS is based on the merge sort algorithm, extended to calculate the dominance set for each solution. MNDS achieves a best computational complexity of $\Theta(NlogN)$, while the worst case is $\Theta(MN^2)$, where $N$ corresponds to the population size and $M$ is the number of objectives.

The rest of the paper is organized as follows. Section~\ref{sec:background} briefly presents current works aiming to reduce the computational cost of non-dominated sorting. Section~\ref{sec:algorithm} describes our proposal in detail. Experimental work and results are provided in Section~\ref{sec:experimentalWork}. Finally, Section~\ref{sec:conclusions} highlights the conclusions and outlines future work.
 
\section{Background and Related Work}
\label{sec:background}

Non-dominated sorting is based on the concept of Pareto-dominance between vectors (or solutions, in the context of EAs). 
Let $P$ be a population of $N$ solutions, $\{s_1,\ldots,s_N\} \in P$, where each solution contains a vector of $M$ objectives to minimize, $(s_{i_1},\ldots, s_{i_M}), \forall i \in\{1,\ldots,N\}$. A solution $s_i$ dominates a solution $s_j$, denoted by $s_i \preceq s_j$, if the vector of objectives of $s_i$ is partially less than the vector of objectives of $s_j$, i.e., $\forall m \in \{1,\ldots,M\}, s_{i_m}\leq s_{j_m} \wedge \exists m' \in \{1,\ldots,M\}\ s.t.\ s_{i_{m'}}<s_{j_{m'}}$ (we assume minimization without loss of generality).
Given a set of solutions, those solutions which are non-dominated by any other are assigned rank 1. If these solutions are removed, then those solutions which are non-dominated by any other are assigned rank 2, and so on.

Kung et al~\cite{Kung} were the first to propose a method based on the divide-and-conquer idea to find maximal elements of a set of vectors, paving the way for further studies. Reducing the complexity of the non-dominated sorting is a matter of active research. The original implementation of NSGA (Non-dominated Sorting Genetic Algorithm)~\cite{Srinivas1994} had a complexity of $\Theta(MN^3)$. A later version, in NSGA-II~\cite{Deb2002}, the \emph{Fast Non-dominated Sorting} reduced the cost to $\Theta(MN^2)$. 
Table~\ref{table:algoritmos} shows both computational and spatial costs of the most representative algorithms for non-dominated sorting and how they compare against the two variants of our current proposal (MNDS).

\begin{table}
\centering
\caption{Complexity of non-dominated sorting algorithms representative of the state-of-the-art.}
\label{table:algoritmos}

\begin{tabular}{cccc}
\hline
  \multirow{2}{*}{Algorithm} 
        & \multicolumn{3}{c}{Complexity} \\ \cline{2-4}
     &Best Case & Worst Case& Space \\  
\hline\hline
FNDS~\cite{Deb2002} & $MN^2$ & $MN^2$ & $N^2$  \\
Dominance Tree~\cite{DominanceTree} & $MNlogN$ & $MN^2$ & $M$  \\ \hline
Deductive Sort~\cite{DeductiveSort} & $MN\sqrt[]{N}$ & $MN^2$ & $N$  \\ 
Corner Sort~\cite{CornerSort} & $MN\sqrt[]{N}$ & $MN^2$ & $N$  \\      \hline
ENS-SS~\cite{ENS} & $MN\sqrt[]{N}$ & $MN^2$ & $1$  \\      
ENS-BS~\cite{ENS} & $MNlogN$ & $MN^2$ & $1$  \\ \hline
\multirow{2}{*}{ENS-NDT~\cite{ENS-NDT}} & {$MNlogN$ if $M>logN$}   & \multirow{2}{*}{$MN^2$}& \multirow{2}{*}{$NlogN$} \\ & $Nlog^2N$ &  & \\ \hline
M-Front~\cite{MFront} & $MN$ & $MN^2$ & $MN^2$\\
DDA-NS~\cite{DDA-NS} & $MN^2$ & $MN^2$ & $N^2$  \\ \hline
HNDS~\cite{HNDS} & $MN\sqrt[]{N}$ & $MN^2$ & $N$  \\
BOS~\cite{BOS} & $MNlogN$ & $ MNlogN+MN^2  \equiv MN^2$ & $N^2$  \\    
\hline\hline
MNDS & $2NlogN \equiv NlogN $  &  $MNlogN+(M-2)N^2 \equiv MN^2$ & $N^2$  \\
\hline
\end{tabular}
\end{table}

We briefly summarize next the different strategies used by each of these algorithms (they are fully described in the provided references):

\begin{itemize}
	\item \emph{Fast Non-dominated Sorting (FNDS)}~\cite{Deb2002} compares each solution with the rest of the solutions of the population to obtain their dominance relationship. While carrying out this comparison, each solution stores those solutions that it dominates in a list. Once the comparisons are done, the lists of dominated solutions are traversed to rank them.
	\item \emph{Dominance Tree}~\cite{DominanceTree} uses a divide-and-conquer strategy to obtain the dominance relationships among the population solutions. These relationships are stored in a tree-like data structure called dominance tree.
	\item \emph{Deductive Sort}~\cite{DeductiveSort} iterates through the population repeatedly, comparing the solutions one by one. Non-dominated solutions are assigned to the corresponding rank and eliminated from the population.
	\item \emph{Corner Sort}~\cite{CornerSort} reduces the number of comparisons using two strategies: (i) as \emph{Deductive Sort}, it avoids comparing solutions marked as dominated; the second strategy (ii) shows a preference for comparing corner solutions when determining the dominance between solutions.
	\item \emph{Efficient Non-dominated Sort (ENS)}~\cite{ENS} calculates the rank of each solution at a time. To do so, it sorts the first objective using the lexicographical comparison\footnote{\label{lexOrder}The lexicographical comparison between two solutions compares the value of the objectives of both solutions starting the first one. If the values are the same, then the second objectives are considered. This is carried out iteratively until the values are different or their objectives are exactly the same.}. Then, it looks for the rank of each solution using a sequential search strategy (version ENS-SS) or a binary search (version ENS-BS).
	\item \emph{M-Front}~\cite{MFront} proposes to modify the typical MOEA's structure to improve they performance. In order to reduce the number of comparisons among solutions, the M-Front algorithm applies the geometric and algebraic properties of the Pareto dominance to perform interval queries using a nearest neighbor search. M-Front defines a special data structure named \textit{archive} where all non-dominated individuals are stored. Additionally, M-Front stores all solutions in lists and uses a K-d tree for nearest neighbor search.
	\item \emph{Hierarchical Non-Dominated Sorting (HNDS)}~\cite{HNDS} minimizes the number of comparisons of objectives by ordering the population by the first objective and then by comparing the first solution with the rest of the solutions. These solutions are moved to an auxiliary list if they are not dominated by the first solution or a list of dominated solutions otherwise. The first solution is assigned to its corresponding rank and then the algorithm iterates until all the solutions are assigned their corresponding rank.
    \item \emph{Dominance Degree Approach for Non-dominated Sorting (DDA-NS)}~\cite{DDA-NS} is based on the concept of dominance matrix to build their \textit{dominance degree matrix} to be applied to compute the ranking of each solution.
	\item \emph{Best Order Sort (BOS)}~\cite{BOS} sorts the population by each objective, resolving ties by means of lexicographical comparison. For each objective and solution $s_i$, it searches those solutions that are not worse than $s_i$. These solutions are stored in a set $T$ associated with $s_i$. BOS will look at $T$ for the $s_j$ solution with the worst rank $r$. The rank of $s_i$ will be $r+1$.    
    \item \emph{Efficient Non-Dominated Sort with Non-Dominated Tree (ENS-NDT)}~\cite{ENS}-NDT extends the ENS-BS~\cite{ENS} algorithm using a new data structure, a variant of a bucket k-d tree, named \emph{Non-Dominated Tree (NDTree)}. ENS-NDT is similar to ENS-BS but in the binary search it uses a NDTree instead of an array to store the fronts, speeding up the domination checking.
 
\end{itemize}

\section{Merge Non-Dominated Sorting}
\label{sec:algorithm}

Merge Non-Dominated Sorting (MNDS) adapts the \emph{merge sort} algorithm~\cite{Sant04} to the non-dominated sorting problem. As it usually happens with these kinds of algorithms, there is a time vs. memory trade-off. In our algorithm the storage of the \textit{dominance set} of each solution allows a reduction of the computational time. Given a solution $s_i \in P$, its \textit{dominance set} contains all solutions that dominate $s_i$, i.e., each $s_j \in P$ such that $s_j \preceq s_i$. 

The overall idea behind our proposal is based on obtaining the dominance set of each solution of the population, and then calculating their ranking based on their corresponding dominance set. To obtain the dominance set for each solution $s_i$, it is necessary to sequentially order the population, by each of the objectives. The order output by the $m$-th objective is the input for objective $m+1$-th. A key point of our approach is the treatment of ties. In the case of the first objective, a lexicographical\footnotemark[\getrefnumber{lexOrder}] comparison is used to break ties. If there are ties for all objective values, the second solution is considered a duplicate. For the rest of objectives, ties are broken using the order of the population output by the previous objective. This is automatically done by the merge sort\footnote{Merge sort is a stable sorting algorithm: when it rearranges the population and there is a tie between two solutions, the relative position of both solutions in the population is maintained.} algorithm.

We next formalise this approach and provide an step-by-step example to illustrate our algorithm.

In a population $P$, where each solution contains a vector of $M$ objective values, the dominance set of solutions $s_i \in P$ can be obtained by sorting $P$ iteratively by each objective as follows:

\begin{itemize}
  \item For the first objective ($m=1$), the individuals are ordered taking into account the objective function value of the objective. In case of ties a lexicographical order is used to rank the individuals. The dominance set of the $i$-th ordered solution $s_i$, in this order, is composed of the preceding solutions in the ordering $s_i.ds = \{ s_1,\ldots,s_{i-1} \}$.
  \item For the remaining objectives $(1< m \leq M)$, we sort the population (previously sorted by objective $m-1$) by each objective $m$. In case of a tie, both solutions maintain the order from the previous ($m-1$ objective) sorting. The dominance set of the $i$-th solution $s_i$, in this order, is composed of the preceding solutions in the $m$-ordering, intersected with the previous dominance set of that solution: $s_i.ds = s_i.ds \cap \{s_1,\ldots,s_{i-1}\}$.
\end{itemize}

After sorting by the last objective, the dominance set of each solution, $s_i.ds$, contains all the solutions that dominate $s_i$. The rank of a solution $s_i \in P$ will be the next rank to the worst rank of all the solutions $s_j \in s_i.ds$. In case that $s_i.ds$ is empty, $s_i$ is assigned rank 1.

\begin{table}
\centering
\caption{Example Population}
\label{table:example}
\setlength\tabcolsep{3.3pt}
{\begin{tabular}{cc||cl|cl|cl||cc}
\hline
\multicolumn{2}{c}{Population$\rightarrow$} & \multicolumn{2}{c}{Order obj. 1$\rightarrow$}& \multicolumn{2}{c}{Order obj. 2$\rightarrow$}& \multicolumn{2}{c}{Order obj. 3$\rightarrow$}& \multicolumn{2}{c}{Rank}  
\\ 
\hline
 id & obj. values& id & dom. set & id & dom. set & id & dom. set & id &R.\\  \hline
1 & 34,30,40 & 4  & $\emptyset$ & 1  & $\emptyset$ & 2  & $\emptyset$ & 1  & 1\\
2 & 33,34,30 & 3  & 4           & 5  & 1           & 3  & $\emptyset$ & 2  & 1\\
3 & 32,32,31 & 2  & 3,4         & 8  & 1,5         & 12 & 2,3         & 3  & 1\\
4 & 31,34,34 & 14 & \textit{dup. sol. 2} & 3  & $\emptyset$ & 7  & 3           & 4  & 1\\
5 & 34,30,41 & 1 &  2..4        & 7  & 1,3,5,8     & 4  & $\emptyset$ & 5  & 2\\
6 & 36,35,36 & 5  & 1..4        & 4  & $\emptyset$ & 6  & 2..4,7      & 6  & 3\\
7 & 36,33,32 & 8  & 1..5        & 2  & 3,4         & 11 & 2..4,6,7    & 7  & 2\\
8 & 35,31,43 & 10 & 1..5,8      & 10 & 1..5,8      & 10 & 2,3         & 8  & 2\\
9 & 37,36,39 & 7  & 1..5,8,10   & 6  & 1..5,7,8,10 & 9  & 2..4,6,7    & 9  & 3\\
10& 35,34,38 & 6  & 1..5,7,8,10 & 9  & 1..8,10     & 1  & $\emptyset$ & 10 & 2\\
11& 38,38,37 & 9  & 1..8,10     & 12 & 1..10       & 5  & 1           & 11 & 3\\
12& 39,37,31 & 13 & \textit{dup. sol. 9} & 11 & 1..10 & 8  & 1,5         & 12 & 2\\
13& 37,36,39 & 11 & 1..10       &   &              &    &             & 13 & 3\\
14& 33,34,30 & 12 & 1..11       &   &              &    &             & 14 & 1\\
 \hline
 \end{tabular}}
\end{table}

\begin{figure}
  \centering
  \vspace{-65pt}
  \scriptsize {\input{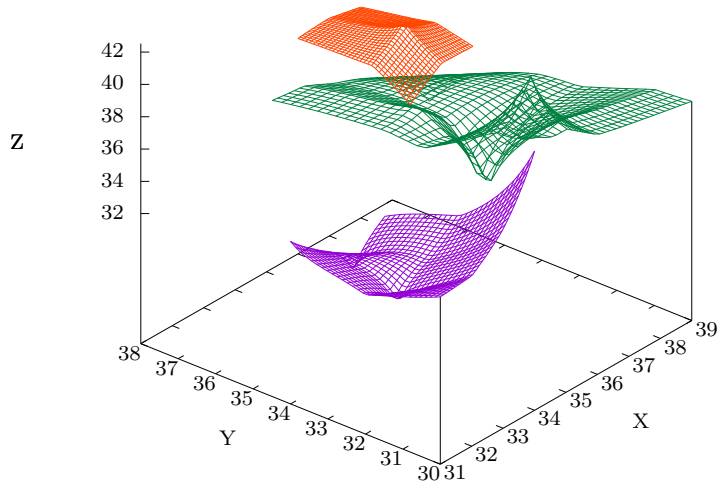}}
  \vspace{-20pt}
  \caption{3D representation of the sample population}
  \label{fig:example}
\end{figure}

To illustrate the working of MNDS, we include in 
Table~\ref{table:example} an step-by-step example using a population $P$ composed of 14 solutions and 3 objectives. The first two columns show the identifier of each solution and the values of each objective. The following three pairs of columns show the solution identifier and the content of the dominance set after being ordered by the first, second and third objectives respectively.

When sorting by the first objective, the pairs $\{s_1,s_5\}$ and $\{s_7,s_6\}$ have the same objective function value. Comparing both pairs lexicographically, we obtain $s_1\preceq s_5$ and $s_7\preceq s_6$. As merge sort is a stable sorting algorithm, the relationship between $\{s_1,s_5\}$ and $\{s_7,s_6\}$ will be maintained, in case of ties, when ordering by the next objectives. The solutions 13 and 14 are duplicates of solutions 9 and 2, so they are removed from the population while carrying out the sorting. They will be added back again to the population after obtaining the ranking (algorithms such as NSGA-II need to keep their population size fixed).

Following the steps previously described the dominance set of \textit{solution} 7, for example, is obtained as follows:
\begin{itemize}
  \item Objective 1: The population is sorted (from lowest to highest) using this first objective. Note that solution 7 is ahead of solution 6 due to lexicographical comparison\footnote{Note that $s_7$ has the same objective value for the first objective than $s_6$, however, $s_7$ is ahead $s_6$}. Once the population is sorted by objective 1, solutions 4, 3, 2, 1, 5, 8 and 10, are smaller (dominate in this objective) than solution 7, i.e., $s_7.ds =\{1,2,3,4,5,8,10\}$
  \item Objective 2: After sorting the population by objective 2, solutions 1, 5, 8 and 3 dominate the solution 7 in objective 2. Therefore, $s_7.ds=s_7.ds\cap\{1,3,5,8\}=\{1,2,3,4,5,8,10\}\cap\{1,3,5,8\}=\{1,3,5,8\}$
  \item Objective 3: After sorting the population by objective 3, solutions 2, 3 and 12 are ranked before than solution 7. Therefore, $s_7.ds=s_7.ds\cap \{2,3,12\}=\{1,3,5,8\}\cap\{2,3,12\}=\{3\}$
\end{itemize}

Finally, the ranks of the solutions are obtained based on the dominance sets and duplicates are inserted again with their corresponding rank. Figure~\ref{fig:example} shows a 3D representation of the example points in Table~\ref{table:example}. The three planes correspond to the three ranks. The higher the plane the higher the rank, having the lowest plane the first rank value.

\subsection{Formalization of the MNDS Algorithm}
As it can be observed in Algorithm~\ref{alg:mnds}, MNDS receives the population to sort as the only parameter. The process followed by MNDS can be divided into the following three phases:

\begin{enumerate}
  \item Sort the population by the \textit{first objective} and create the dominance set of each solution. Ties are broken using lexicographical comparison and the duplicated solutions are moved to a list of $duplicates$ solutions (Algorithm~\ref{alg:mnds} line~\ref{algLn:mnds:dup}). This list is composed of tuples \textit{(duplicate solution, original solution)}. It is worth noting that although this requires more memory than just keeping the original solution with a list of duplicates, our solution is faster as it avoids searching through such list. The rank of duplicate solutions is assigned at the end of the algorithm (line~\ref{algLn:mnds:updateRanking}).
  
  \item Iteratively sort the population by the rest of the objectives $1~<~m~\leq~M$. In case that in any iteration all the dominance sets are empty, i.e. there is no dominance, this phase ends and the third phase is omitted as all solutions belong to the first rank.

  \item Calculate the rank of each solution.
\end{enumerate}

Lines~\ref{algLn:mnds:sort1} and~\ref{algLn:mnds:sortrest} in Algorithm~\ref{alg:mnds} correspond to the first and second phases respectively. These phases are in turn further described in Algorithms~\ref{alg:mnds_firstObj} and~\ref{alg:mnds_restOfObjectives}. The calculation of the ranking of each solution (phase 3) corresponds to lines~\ref{algLn:mnds:getRanking} and~\ref{algLn:mnds:updateRanking}. Also, Algorithm~\ref{alg:mnds_restOfObjectives} returns \textit{true} if there is a dominance among the solutions, or \textit{false} otherwise.

\begin{algorithm}
\caption{Merge Non-Dominated Sorting($P$)}
\label{alg:mnds}
\begin{algorithmic}[1]
\Require population $P$
\Ensure  ranking for each solution $R$
    \State $R \leftarrow \emptyset$
    \State $duplicates \leftarrow \label{algLn:mnds:dup} \textbf{SortFirstObjective}(P)$ \label{algLn:mnds:sort1} 
    \If{$\textbf{SortRestOfObjectives}(P)$} \label{algLn:mnds:sortrest} 
      \State $R \leftarrow$ \textbf{GetRanking}($P$) \label{algLn:mnds:getRanking}
	   \State Update the rank of each $duplicates$ solution with the rank of its original solution \label{algLn:mnds:updateRanking}
    \EndIf
    \State $\textbf{return}~R$ 
\end{algorithmic}                 
\end{algorithm}

Algorithms~\ref{alg:mnds_firstObj} and~\ref{alg:mnds_restOfObjectives} sort the population $P$ by the objective~$O$ using Algorithm~\ref{alg:mnds_sort}, $Sort(P,O)$. As previously stated, this algorithm is based on the merge sort algorithm. When sorting by the first objective ($O=1$), in case of ties, the lexicographical comparison is applied (see Algorithm~\ref{alg:mnds_sort} line~\ref{algLn:mnds_sort:msortL1}).

The method $SortFirstObjective(P)$ shown in Algorithm~\ref{alg:mnds_firstObj} implements the sorting by the first objective (phase 1). Line~\ref{algLn:mnds_firstObj:sort} sorts the population $P$ by its first objective using the lexicographic rule in case of ties. 
Next, the loop (from lines~\ref{algLn:mndsFirstObj:for} to~\ref{algLn:mndsFirstObj:endfor}) calculates the dominance set of each solution (lines~\ref{algLn:mnds_firstObj:dsInitialization},~\ref{algLn:mnds_firstObj:incset}) and moves the duplicate solutions (see lines~\ref{algLn:mnds_firstObj:dup1},~\ref{algLn:mnds_firstObj:dup2}) to the $duplicates$ list. Note that, in each iteration $i$, the auxiliary set $incSet$ contains the solutions that dominate the $s_i$ solution.

\begin{algorithm}
\caption{SortFirstObjective($P$)}
\label{alg:mnds_firstObj}
\begin{algorithmic}[1]
\Require population $P$ 
\Ensure population $P$, duplicate solutions $duplicates$
\State $incSet \leftarrow \emptyset$ \Comment{auxiliary incremental set}
\State $duplicates \leftarrow \emptyset$
\State $\textbf{Sort}(P,1)$ \label{algLn:mnds_firstObj:sort}
\State $u \leftarrow P[1]$ \Comment auxiliary solution $u$
\State $incSet \leftarrow incSet \cup u$
\For{$s:P$} \Comment{ $ s_i \in P, \forall i \in \{2,\ldots,|P|\}$} \label{algLn:mndsFirstObj:for} 
	\If{$s \neq u$}
        \State $s.ds \leftarrow incSet$ \label{algLn:mnds_firstObj:dsInitialization}
        \State $incSet \leftarrow incSet \cup s$ \label{algLn:mnds_firstObj:incset}
	\Else
      	\State $duplicates \leftarrow duplicates\cup s$ \label{algLn:mnds_firstObj:dup1}
        \State $P \leftarrow P - s$ \label{algLn:mnds_firstObj:dup2}
    \EndIf
    \State $u \leftarrow s$
\EndFor \label{algLn:mndsFirstObj:endfor}
\State $\textbf{return}~P, duplicates$
\end{algorithmic}                 
\end{algorithm}

The method $SortRestOfObjectives(P)$ shown in Algorithm~\ref{alg:mnds_restOfObjectives} implements the second phase. The first loop iterates through all objectives except the first one. The calculation of the dominance sets is carried out by the internal loop (lines~\ref{algLn:mnds_restOfObjectives:for} to~\ref{algLn:mnds_restOfObjectives:endfor}), which also evaluates if there is dominance among the solutions (line~\ref{algLn:mnds_restOfObjectives:ifdom}). When there is no further dominance, the method ends.

\begin{algorithm}
\caption{SortRestOfObjectives($P$)}
\label{alg:mnds_restOfObjectives}
\begin{algorithmic}[1]
\Require population $P$ 
	\Ensure population $P$, \textit{hasDominance} Boolean with whether there is dominance
	\State $obj \leftarrow 2$
    \State $incSet \leftarrow \emptyset$ \Comment{auxiliary incremental set}
    \State $hasDominance \leftarrow \textbf{true}$
    \While{$obj \leq M \land hasDominance$}     
      \If{$\textbf{Sort}(P, obj)$} 
       \State $hasDominance \leftarrow \textbf{false}$
       \State $incSet \leftarrow \emptyset$
 			\For {$s:P$} \label{algLn:mnds_restOfObjectives:for}
            \State $s.ds \leftarrow s.ds \cap incSet$ \label{algLn:mnds_restOfObjectives:dsUpdate}
              \State $incSet \leftarrow incSet \cup s$ \label{algLn:mnds_restOfObjectives:incset}
              \If{$s.ds \neq \emptyset$} \label{algLn:mnds_restOfObjectives:ifdom}
                  \State $hasDominance \leftarrow \textbf{true}$ \label{algLn:mnds_restOfObjectives:domTrue}
              \EndIf
        \EndFor \label{algLn:mnds_restOfObjectives:endfor}
        \EndIf
		\State $obj \leftarrow obj+1$
    \EndWhile
\State $\textbf{return}~P, hasDominance$
\end{algorithmic}                 
\end{algorithm}

The last phase, the calculation of the population ranking, is implemented by the method $GetRanking(P)$. In this method, the variable $maxRank$ always contains the highest rank value of all evaluated solutions. It worth noting that the rank of a solution $s$ is always in the range $[1, maxRank+1]$. The internal loop (lines~\ref{algLn:GetRanking:for} to~\ref{algLn:GetRanking:endfor}) traverses the dominance set $s.ds$, obtaining the rank ($R[u]$, line~\ref{algLn:GetRanking:ifR}) of each solution in the current dominance set. If that value is greater than current $rank$, the $rank$ value is increased to $R[u]+1$ (line~\ref{algLn:GetRanking:rank}). Likewise, if the value of the $rank$ variable is greater than $maxRank$ (line~\ref{algLn:GetRanking:rankfound}), the $rank$ value is assigned to $maxRank$ and the search ends.

\begin{algorithm}
\caption{GetRanking($P$)}\label{alg:GetRanking}
\begin{algorithmic}[1]
\Require Population $P$  
\Ensure Population Ranking $R$
	\State $R \leftarrow \emptyset$
    \State $maxRank \leftarrow 0$
	\For {$s:P$}
      \State $rank \leftarrow 0$
      \For{$u:s.ds$} \Comment{for each solution $u$ that dominates $s$}\label{algLn:GetRanking:for}
          \If{$R[$u$] \geq rank$}\label{algLn:GetRanking:ifR}
              \State $rank = R[$u$]+1$\label{algLn:GetRanking:rank}
          \EndIf
          \If{$rank > maxRank $}\label{algLn:GetRanking:rankfound}
              \State $maxRank \leftarrow rank$
              \State $R[s]\leftarrow rank$
              \State $\textbf{break}$
          \EndIf
      \EndFor \label{algLn:GetRanking:endfor}
    \EndFor
    \State $\textbf{return}$ $R$
\end{algorithmic}                 
\end{algorithm}

\begin{algorithm}
\caption{Sort($P, O$)}
\label{alg:mnds_sort}
\begin{algorithmic}[1]
\Require population $P$, objective $O$  
\Ensure population $P$ sorted by objective $O$, \newline
$isSorted$ Boolean with whether input $P$ is already ordered
	\State $isSorted \leftarrow \textbf{false}$
    \If{$O>1$}
    \State $isSorted \leftarrow$Sort $P$ using merge sort\label{algLn:mnds_sort:msort1}
    \Else
     \State $isSorted \leftarrow$ Sort $P$ using merge sort applying lexicographical order in case of ties \label{algLn:mnds_sort:msortL1}
	\EndIf
\State $\textbf{return}~P$, $isSorted$
\end{algorithmic}                 
\end{algorithm}

\subsection{Computational and Spatial Complexity }
\label{subsec:complexity}

MNDS is based on the merge sort algorithm (see Algorithm~\ref{alg:mnds_sort}) which has a best and worst complexity of $\Theta(NlogN)$. The complexity of MNDS (Algorithm~\ref{alg:mnds}), in the worst case scenario, is the sum of the complexities of the methods shown in Algorithms~\ref{alg:mnds_firstObj},~\ref{alg:mnds_restOfObjectives} and~\ref{algLn:mnds:getRanking}, which are calculated as follows: 
\begin{itemize}
\item Algorithm~\ref{alg:mnds_firstObj}: The worst case belongs to the sorting which has a complexity of $\Theta(NlogN)$.
\item Algorithm~\ref{alg:mnds_restOfObjectives}: The inner loop calculates the dominance of all solutions in $P$ ($\Theta(N)$), which computes the intersection $s.ds \cap incSet$ ($\Theta(N)$) so this loop has a complexity of ($\Theta(N^2)$). The external loop sorts ($\Theta(NlogN)$) the population $P$ for each objective except the first one, i.e., objectives 2 to $M$. Therefore, the complexity for this algorithm is $\Theta((M-1)(NlogN+N^2))$
\item Algorithm~\ref{algLn:mnds:getRanking} is composed of two nested loops, so its complexity is $\Theta(N^2)$.
\end{itemize}
Therefore, the complexity of MNDS is the sum of $\Theta(NlogN)$, $\Theta((M-1)(NlogN+N^2))$, and $\Theta(N^2)$ which equals to $\Theta(MN^2)$.

The best case occurs when there is no dominance among the solutions and in the first iteration of the Algorithm~\ref{alg:mnds_restOfObjectives} all the \textit{dominance sets} of the solutions are empty. In this case, it is not necessary to compute intersections in the Algorithm~\ref{alg:mnds_restOfObjectives} because the sets are disjoints and the complexity is $\Theta(NlogN)$. 
 It is worth noting that this algorithm improves its complexity towards $\Theta(NlogN)$ when the number of fronts decreases, which typically happens as number of iterations evolves with generations. In the rest of the algorithms analyzed, their complexity increases towards $\Theta(MN^2)$ as the number of the fronts decreases.

The spatial complexity is determined by the size of the \textit{dominance sets} of each solution ($|s_i.ds|=|P|$) which corresponds to $\Theta(N^2)$.

\section{Experimental Work}
\label{sec:experimentalWork}

\subsection{Implementation details}

To validate the performance of MNDS, we compare MNDS with four well-known algorithms from the literature: BOS~\cite{BOS}, HNDS~\cite{HNDS}, ENS-SS and ENS-BS~\cite{ENS}. To do so, we use the BOS implementation provided by the authors\footnote{\label{githubBOS}\texttt{\url{https://github.com/Proteek/Best-Order-Sort}}}. For the ENS-SS and ENS-BS~\cite{ENS} algorithms, the implementations provided by Buzdalov\footnote{\label{githubMBuzdalov}\texttt{\url{https://github.com/mbuzdalov/non-dominated-sorting}}} were used with minimal modifications.
In addition, we also implemented the HNDS~\cite{HNDS} algorithm from scratch. We needed to make two changes to the description provided in the author's paper because the results obtained were erroneous in the case where we get more than one solution with one or more equal objective values.
The number of comparisons is shown in Table~\ref{table:comparisons} and corresponds to the number of times that each algorithm compares the value of an objective between two solutions.

In relation to MNDS\footnote{MNDS will be included into future versions of the jMetal framework including the source code.} implementation, we make use of \textit{bitsets} to deal with sets operations. The motivation behind using bitsets to represent sets is their capability of maintaining the set sorted to facilitate the insertion of elements with a complexity of $\Theta(1)$ while in other implementations such as lists, their cost is $\Theta(log N)$. It is worth noting that we use \emph{sorted} sets to speed up the intersection operation between sets. Furthermore, our implementation of bitsets considers the range of the values in the set $[min, max]$, the intersection between two sets $a$ and $b$ is only applied within the range $[Max(a.min, b.min), Min(a.max, b.max)]$.

\subsection{Experimental settings}

In order to compare the algorithms previously discussed, three types of experiments were carried out:

\begin{enumerate}[a)]

  \item Varying the number of objectives for a fixed number of solutions, using the BOS dataset\footnotemark[\getrefnumber{githubBOS}].

  \item Varying the population size for a fixed number of objectives, using again the BOS dataset.

  \item Varying the number of objectives for a fixed population size, using datasets generated by NSGA-II. In this case, we have additionally obtained the number of comparisons made by each algorithm.

\end{enumerate}

The original BOS dataset contains 10,000 solutions with up to 10 objectives; we extended it to 20 objectives, generating the new values randomly. 

In the experiment a), the algorithms were executed varying the number of objectives between 3 and 20, with population sizes of 500, 1,000, 5,000. 
In b), the size of the population ranged between 500 and 10,000 with an increase of 1,000 for 5, 10, 15 and 20 objectives. 

Finally, for the experiment c), the NSGA-II~\cite{Deb2002} implementation from jMetal~\cite{jMetal} was used to generate 16 datasets obtained after 200 generations for the DTLZ1~\cite{DTLZ}, DTLZ2~\cite{DTLZ}, WFG1~\cite{WFG} and WFG2~\cite{WFG}  problems with 5, 10, 15 and 20 objectives. The population size used was 800 solutions. All the algorithms were executed 5,000 times under the same conditions using the execution time as performance measure. The final execution time was calculated averaging those 5,000 executions.

It worth noting that the NSGA-II was used with the same problems and applying the same configuration as the one defined in the paper describing BOS. In this way, when comparing MNDS against BOS, we are also comparing MNDS, indirectly, with the other algorithms that were also compared with the BOS algorithm, i.e., fast non-dominated sorting, deductive sort, corner sort and divide-and-conquer sort.

The computer and software versions used have the following features:
\begin{itemize}
  \item Windows Server 2016 Standard.
  \item 8 Quad-Core AMD Opteron Processor 8356 2.29GHz.
  \item 256GB of RAM Memory.
  \item Java version: 1.8.0-121, 64 bits.
\end{itemize}

\subsection{Results}
\label{sec:results}

\begin{figure*}
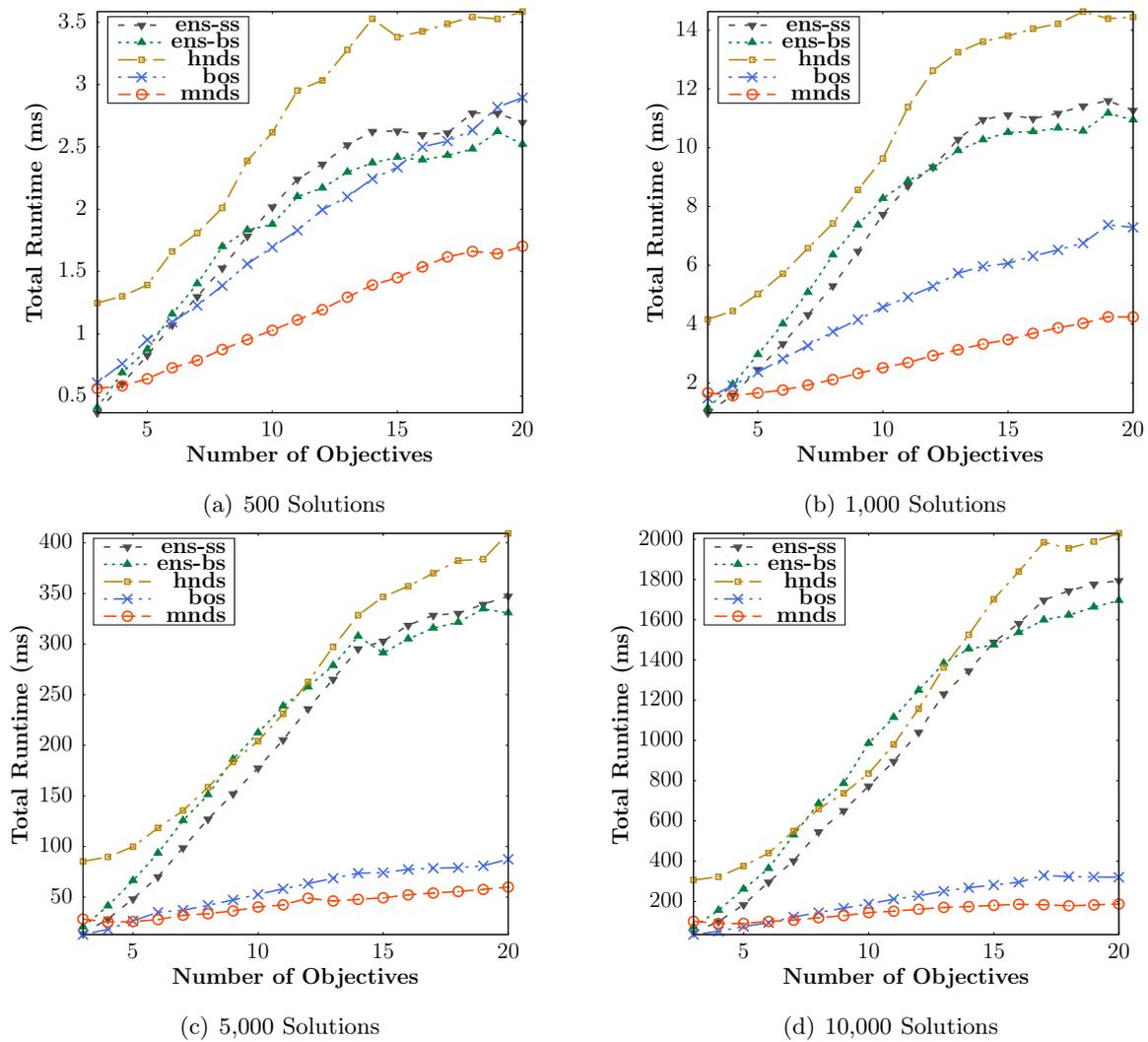
 
    \centering
\Large{\textbf{
\subfigure[500 Solutions]{
  		\resizebox{0.6\width}{!}{\input{RT500.tex}}
  		\label{fig:500sol}
  	}\hspace{0.1\textwidth}
    \subfigure[1,000 Solutions]{
  		\resizebox{0.6\width}{!}{\input{RT1000.tex}}
  		\label{fig:1000sol}
  	}\hspace{0.1\textwidth}
    \subfigure[5,000 Solutions]{
  		\resizebox{0.6\width}{!}{\input{RT5000.tex}}
  		\label{fig:5000sol}
  	}\hspace{0.1\textwidth}
    \subfigure[10,000 Solutions]{
  		\resizebox{0.6\width}{!}{\input{RT10000.tex}}
  		\label{fig:10000sol}
  	}
}}
\caption{Experiment 1. Results with a fixed number of solutions using the BOS dataset.}
\label{fig:resultadosBOS_fixedSols}
\end{figure*}

\begin{figure*}
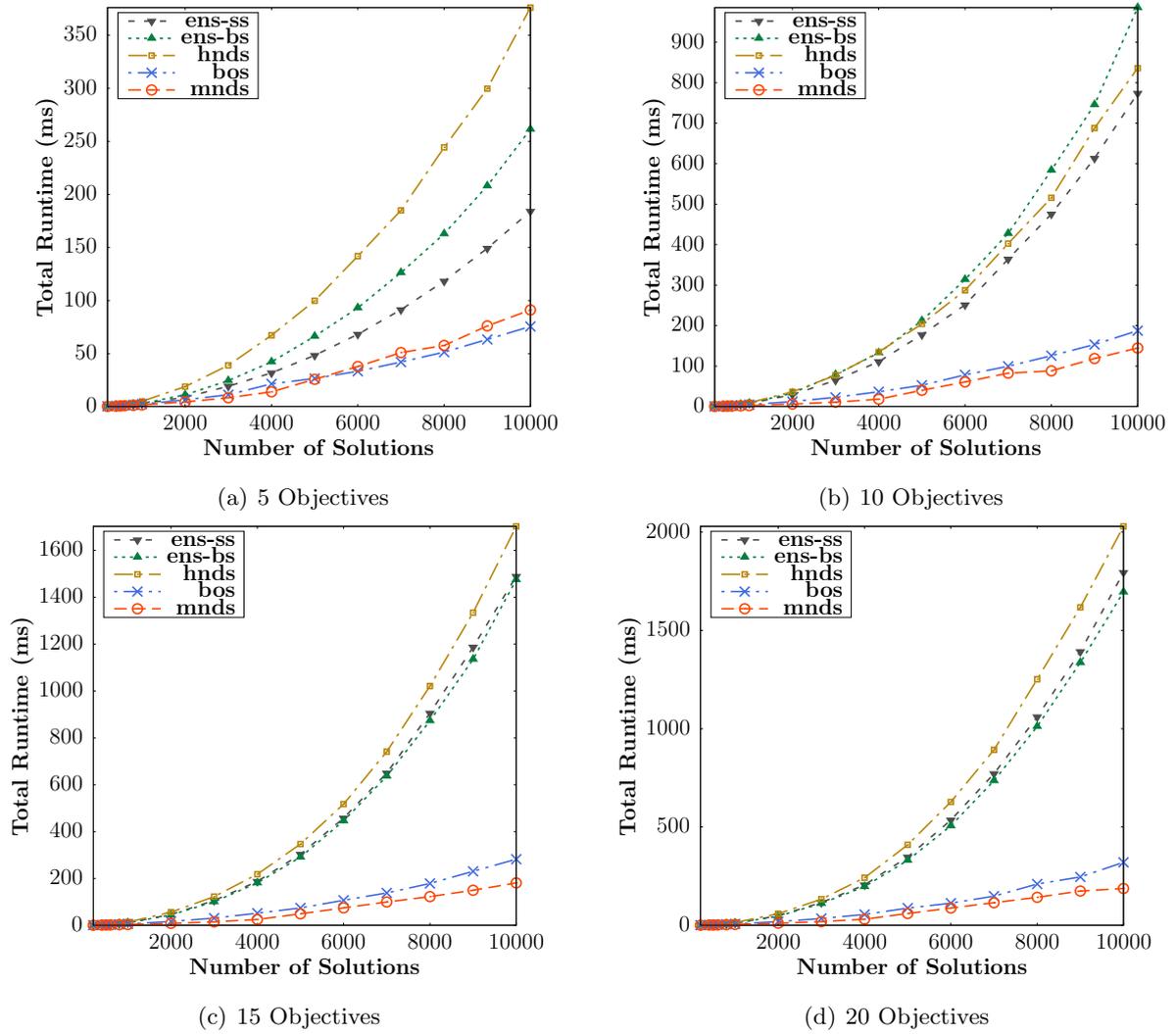
 
    \centering 
\Large{\textbf{
	\subfigure[5 Objectives]{
  		\resizebox{0.6\width}{!}{\input{OBJ5.tex}}
  		\label{fig:5obj}
  	}\hspace{0.1\textwidth}
    \subfigure[10 Objectives]{
  		\resizebox{0.6\width}{!}{\input{OBJ10.tex}}
  		\label{fig:10obj}
  	}\hspace{0.1\textwidth}
    \subfigure[15 Objectives]{
  		\resizebox{0.6\width}{!}{\input{OBJ15.tex}}
  		\label{fig:15obj}
  	}\hspace{0.1\textwidth}
    \subfigure[20 Objectives]{
  		\resizebox{0.6\width}{!}{\input{OBJ20.tex}}
  		\label{fig:20obj}
  	}    
}}
\caption{Experiment 2. Results with a fixed number of objectives using the BOS dataset.}
\label{fig:resultadosBOS_fixedObj}
\end{figure*}

\begin{figure*} 
    \centering
    \Large{\textbf{
	\subfigure[DTLZ1]{
  		\resizebox{0.6\width}{!}{\input{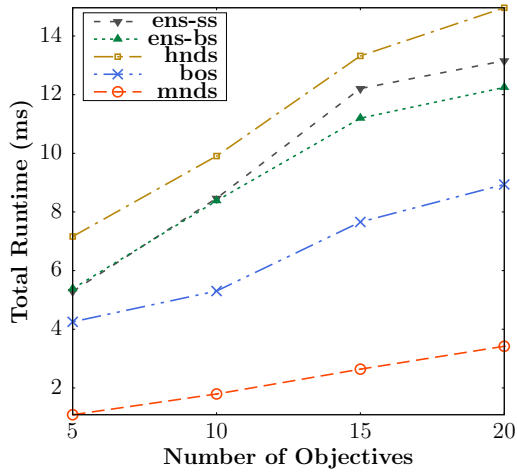}}
  		\label{fig:DTLZ1}
  	}\hspace{0.1\textwidth}
    \subfigure[DTLZ2]{
  		\resizebox{0.6\width}{!}{\input{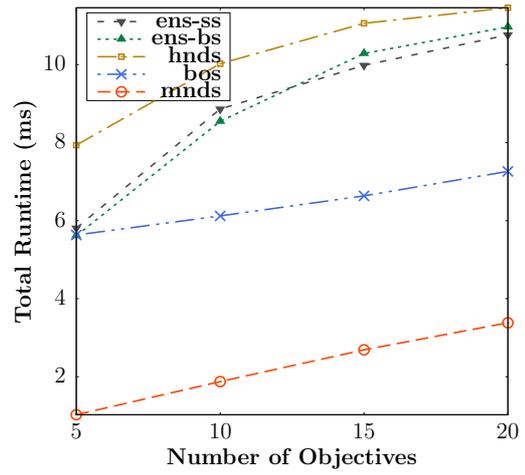}}
  		\label{fig:DTLZ2}
  	}\hspace{0.1\textwidth}
    \subfigure[WFG1]{
  		\resizebox{0.6\width}{!}{\input{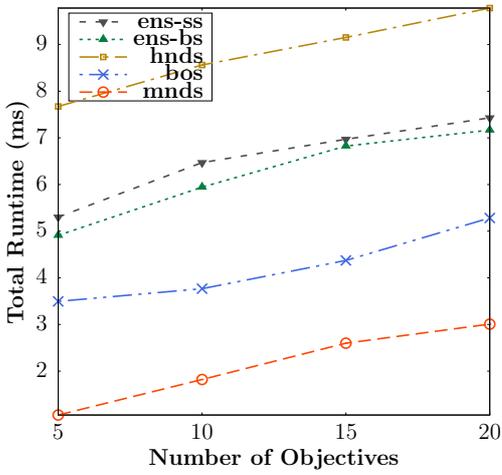}}
  		\label{fig:WFG1}
  	}\hspace{0.1\textwidth}
    \subfigure[WFG2]{
  		\resizebox{0.6\width}{!}{\input{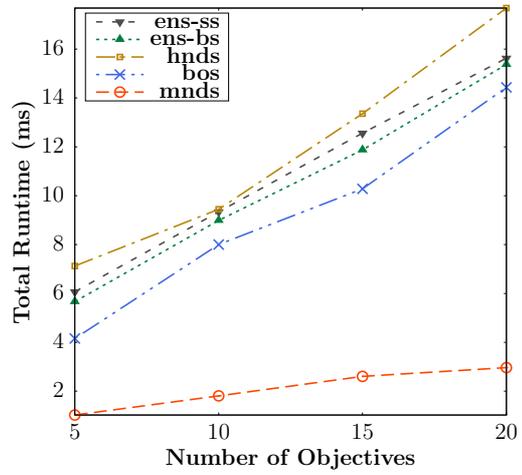}}
  		\label{fig:WFG2}
  	}
    }}
\caption{Experiment 3. Results with the Dataset Generated by NSGA-II.}
\label{fig:resultadosNSGA}
\end{figure*}

The execution times obtained by all the algorithms in the first two experiments are shown in  Figs.~\ref{fig:resultadosBOS_fixedSols} and~\ref{fig:resultadosBOS_fixedObj} respectively. We can observe that MNDS outperforms the rest of the algorithms, except in the case of 5 objectives, where BOS performs slightly better. The MNDS algorithm was designed to work efficiently with large population sizes as well as with a large number of objectives. As a result, the algorithm maintains a very high performance even if we increase the number of objectives or the size of the population. With few objectives ($<$ 5) and/or a small number of solutions (see Figs.~\ref{fig:resultadosBOS_fixedSols}(a) and~\ref{fig:resultadosBOS_fixedSols}(b)), the behavior of all the compared techniques is similar. In those figures we can also observe that (i) as the number of objectives increases or the size of the population increases, the rest of the algorithms suffer a performance degradation; (ii) only the BOS algorithm presents a performance close to MNDS when using the BOS dataset. 
The computing times obtained with the datasets generated by NSGA-II indicate that the differences with the other algorithms are noticeable, as it can be observed in Fig.~\ref{fig:resultadosNSGA}.

Since MNDS is based on the merge sort algorithm, the number of comparisons performed is practically the minimum possible, with the exception of the ordering of the first objective, where the lexicographical organization may require additional comparisons. Table~\ref{table:comparisons} shows how MNDS requires a number of comparisons, in the third experiment c), that is at least one order of magnitude less than the best of the other algorithms compared.

\begin{table}
\centering
\caption{Experiment 3. Number of comparisons made by the algorithms.}
\label{table:comparisons}
\setlength\tabcolsep{6pt}
\begin{tabular}{cclllll|}
\cline{2-7}
\multicolumn{1}{c|}{}           &Obj& \multicolumn{1}{c}{BOS} & \multicolumn{1}{c}{ENS-SS} & \multicolumn{1}{c}{ENS-BS} & \multicolumn{1}{c}{HNDS} & \multicolumn{1}{c|}{MNDS} \\ \hline  
\multicolumn{1}{|c|}{\multirow{4}{*}{\rotatebox[origin=c]{90}{DTLZ1}}} & 5    & 3.72e+08                & 1.22e+06                   & 1.22e+06                   & 1.23e+06                 & \textbf{3.57e+04}                  \\
\multicolumn{1}{|c|}{}                       & 10   & 3.29e+08                & 1.64e+06                   & 1.64e+06                   & 1.65e+06                 & \textbf{6.95e+04}                  \\
\multicolumn{1}{|c|}{}                       & 15   & 4.02e+08                & 2.31e+06                   & 2.31e+06                   & 2.32e+06                 & \textbf{1.02e+05}                  \\
\multicolumn{1}{|c|}{}                       & 20   & 4.24e+08                & 2.71e+06                   & 2.71e+06                   & 2.72e+06                 & \textbf{1.34e+05}                  \\ \hline
\multicolumn{1}{|c|}{\multirow{4}{*}{\rotatebox[origin=c]{90}{DTLZ2}}} & 5    & 4.32e+08                & 1.27e+06                   & 1.27e+06                   & 1.28e+06                 & \textbf{3.71e+04}                  \\
\multicolumn{1}{|c|}{}                       & 10   & 3.71e+08                & 1.74e+06                   & 1.74e+06                   & 1.75e+06                 & \textbf{6.92e+04}                  \\
\multicolumn{1}{|c|}{}                       & 15   & 3.44e+08                & 1.90e+06                   & 1.90e+06                   & 1.91e+06                 & \textbf{1.02e+05}                  \\
\multicolumn{1}{|c|}{}                       & 20   & 3.76e+08                & 2.23e+06                   & 2.23e+06                   & 2.24e+06                 & \textbf{1.34e+05}                  \\ \hline
\multicolumn{1}{|c|}{\multirow{4}{*}{\rotatebox[origin=c]{90}{WFG1}}}  & 5    & 2.54e+08                & 1.14e+06                   & 1.14e+06                   & 1.15e+06                 & \textbf{3.69e+04}                  \\
\multicolumn{1}{|c|}{}                       & 10   & 2.07e+08                & 1.23e+06                   & 1.23e+06                   & 1.24e+06                 & \textbf{6.96e+04}                  \\
\multicolumn{1}{|c|}{}                       & 15   & 2.33e+08                & 1.31e+06                   & 1.31e+06                   & 1.32e+06                 & \textbf{1.02e+05}                  \\
\multicolumn{1}{|c|}{}                       & 20   & 2.63e+08                & 1.40e+06                   & 1.40e+06                   & 1.41e+06                 & \textbf{1.22e+05}                  \\ \hline
\multicolumn{1}{|c|}{\multirow{4}{*}{\rotatebox[origin=c]{90}{WFG2}}}  & 5    & 3.51e+08                & 1.16e+06                   & 1.16e+06                   & 1.18e+06                 & \textbf{3.69e+04}                  \\
\multicolumn{1}{|c|}{}                       & 10   & 5.23e+08                & 1.76e+06                   & 1.76e+06                   & 1.77e+06                 & \textbf{6.94e+04}                  \\
\multicolumn{1}{|c|}{}                       & 15   & 5.93e+08                & 2.54e+06                   & 2.54e+06                   & 2.55e+06                 & \textbf{1.00e+05}                  \\
\multicolumn{1}{|c|}{}                       & 20   & 7.19e+08                & 3.71e+06                   & 3.71e+06                   & 3.72e+06                 & \textbf{1.28e+05}                  \\ \hline
\end{tabular}
\end{table}

\section{Conclusions and Future Work}
\label{sec:conclusions}

In this paper, we have presented a new and efficient algorithm for computing the non-dominated sorting called Merge Non-Dominated Sorting (MNDS) based on the merge sort algorithm. The experimental work showed that MNDS strongly outperforms the current state of the art algorithms in terms of running time.

As future work we plan to improve our approach. Particularly, we think that the algorithm used to calculate the ranking of each solution from the domination sets can be improved by the use of different search methods and data structures. Furthermore, we will also consider the possibility of parallelizing the algorithm. 






\bibliographystyle{plain}
\bibliography{IEEEabrv,references}

\end{document}